\documentclass[11pt]{article}
\pdfoutput=1
\usepackage{eamt12}
\usepackage{times}
\usepackage{latexsym}
\usepackage{mathptmx}
\usepackage[vlined,ruled]{algorithm2e}
\usepackage{misc_defns}
\usepackage{math_defns}
\usepackage{book_defns}
\usepackage{amsmath,amsthm}
\usepackage{amssymb}
\usepackage{graphicx}
\usepackage{microtype}
\usepackage{url}
\usepackage{latexsym}
\title{Exploiting Out-of-Domain Data Sources for Dialectal Arabic Statistical Machine Translation}

\author{Katrin Kirchhoff\\
  Department of Electrical Engineering\\
  University of Washington\\
  Seattle, WA, USA\\
  {\tt kk2@u.washington.edu}  \And
  Bing Zhao\thanks{Work was done while the author was at SRI International.}\\
  LinkedIn\\
  {\tt bingzhao@gmail.com} \AND
  Wen Wang\\
  SRI International\\
  Menlo Park, CA, USA\\
  {\tt wwang@speech.sri.com}}

\date{}

\begin{document}
\maketitle
\begin{abstract}
  Statistical machine translation for dialectal Arabic is
  characterized by a lack of data since data acquisition involves the
  transcription and translation of spoken language. In this study we
  develop techniques for extracting parallel data for one particular
  dialect of Arabic (Iraqi Arabic) from out-of-domain corpora in
  different dialects of Arabic or in Modern Standard Arabic. We
  compare two different data selection strategies (cross-entropy based
  and submodular selection) and demonstrate that a very small but
  highly targeted amount of found data can improve the performance of
  a baseline machine translation system.  We furthermore report on
  preliminary experiments on using automatically translated speech
  data as additional training data.
\end{abstract}

\section{Introduction}

In the Arabic-speaking world, dialectal Arabic (DA) is used
side-by-side with the standard form of the language, Modern Standard
Arabic (MSA).  Whereas the latter is used for written and formal oral
communication (lectures, speeches), DA is used for everyday, casual
communication.  DA is almost never written; exceptions are
transcriptions of spoken language, e.g., in novels, movie scripts,
or in online blogs or forums.  DA and MSA exhibit strong differences at
the lexical, phonological, morphological, and syntactic levels;
furthermore, the dialects themselves form a similarity continuum that
ranges from closely related to mutual unintelligible. An overview of
the main characteristics of DA can be found in \cite{habash10}.

Most natural language processing (NLP) tools that have been developed
for Arabic have been targeted towards MSA, for which large amounts of
written data exist. NLP for DA suffers from a sparsity of tools as
well as data. Work on DA annotation tools includes the development of
morphological analyzers for Arabic dialects 
\cite{habash05,habash12,habash13}, treebanks \cite{maamouri06} and
parsers \cite{chiang06}, unsupervised \cite{duh05} or supervised
\cite{sabbagh12} training of POS taggers for DA, and lexicon
acquisition \cite{duh06}. However, most of these have been targeted to
the Egyptian or Levantine dialects and do not easily generalize to
other dialects. There are a small number of speech and parallel text
corpora for Egyptian, Levantine, and Iraqi DA, primarily available
from the Linguistic Data Consortium (LDC) and the European Language
Resources Association (ELRA). In general, however, spoken language
needs to be recorded and transcribed to produce text data, which
constitutes a bottleneck for the rapid acquisition of new data.

The lack of training data for DA in statistical machine translation
(SMT) has only been addressed in a few previous studies; the standard
approach has been to simply collect more training data by transcribing
and translating DA speech.  \cite{zbih12} compare utilizing
large amounts of MSA data for training and creating a small corpus of DA training
data. They conclude that simply adding large amounts of mismatched
(MSA) training data does not help, whereas even a small amount of
dialectal data is very useful.  Salloum and Habash
\cite{salloum11,salloum13} propose to transform DA to MSA by means of
a combination of statistical processing and hand-coded transformation
rules, and to then apply MT systems for MSA-to-English. Their work was
on Egyptian Arabic, and porting this approach to a different dialect
involves a fair amount of manual effort and dialect expertise. In
\cite{aminian14} the specific problem of out-of-vocabulary words in MT
for DA is addressed by replacing DA words with their MSA equivalents.

In this paper we attempt to enrich available training data for Iraqi
Arabic by automatically identifying IA-English parallel
data in out-of-domain corpora of MSA and other dialects of Arabic.
This procedure is based on the assumption that at least some dialects
will exhibit similarities with IA. Corpora formally described as MSA
may also contain dialectal data at the subsentential level due to
code-switching (mixed use of MSA and DA), which is common among
Arabic speakers. In principle, automatic dialect identification methods \cite{alorifi08,sadat14,zaidan14}
might be used for this purpose; however, these methods are themselves
error-prone and have not been developed for all dialects of Arabic. 
Our approach is to directly select data that is matched to features (n-grams)
extracted from a sample corpus of the dialect of interest. In addition to finding
dialect-matched data, the selected data is also likely to be matched with respect to topic and style.
Two different data selection methods are
investigated, the widely-used cross entropy method of 
\cite{moore10}, and a more recent submodular data selection method
\cite{wei13}. We demonstrate that the performance of SMT systems for
IA can be improved by selecting a very small amount of highly targeted
out-of-domain data.  In addition, we conduct a preliminary
investigation of the possibility of using automatically translated
speech data as SMT training data.

The paper is structured as follows: we first report on previous work
on data selection for SMT (Section \ref{sec:prior}).  We then describe
the submodular technique used in this paper in detail (Section
\ref{sec:submod}). The data is described in Section \ref{sec:data};
experiments are results are presented in Section \ref{sec:exp}. We
provide conclusions in Section \ref{sec:conc}.

\section{Data Selection: Previous Work}
\label{sec:prior}

A currently widely-used data selection method in SMT (which we also
use as a baseline in Section~\ref{sec:exp}) uses the cross-entropy
between two language models~\cite{moore10}, one trained on the test
set of interest, and another trained on generic or
out-of-domain training data. We call this the \emph{cross-entropy
  method}.  This method trains a test-set specific (or in-domain)
language model, $\text{LM}_{\text{in}}$, and a generic (out-of- or
mixed-domain) language model, $\text{LM}_{\text{out}}$.  Each sentence
$x \in V$ in the training data is scored by both language models and is
assigned the log ratio of the language model probabilities as a score:
\begin{align}
m_\text{ce}(x) = \frac{1}{\ell(x)}\log [
\text{Pr}(x|\text{LM}_{\text{in}}) /
\text{Pr}(x|\text{LM}_{\text{out}}) ]
\label{eqn:log_ratio_lm}
\end{align} 
where $\ell(x)$ is the length
of sentence $x$.  Sentences are then ranked in descending order based on
their scores and the top $N$ sentences are chosen.
Various extensions to this method have been proposed. In
\cite{axelrod11} the monolingual selection method is extended to
bilingual corpora. In \cite{duh13}, neural language models are used
instead of backoff language models. Finally, \cite{mediani14} propose
a different method for drawing the out-of-domain sample and the use of
word-association models to improve the data for training the
out-of-domain language model.

The cross-entropy approach ranks each sentence individually, without
reference to other sentences. Thus, no sentence interactions can be modelled, such as
redundancy at the sentential or sub-sentential level.  Moreover, the method does not have a
theoretical performance guarantee.

\section{Submodular Data Selection}
\label{sec:submod}

Submodular functions \cite{ed70,fujishige2005submodular} were first
developed in mathematics, operations research and economics; more
recently, they have been used for a variety of optimization problems
in machine learning as well. For example, they have been applied to
the problems of clustering \cite{nara07}, observation selection
\cite{krause08}, sensor placement \cite{krause11}, or image
segmentation \cite{jegelka11}. Within natural language processing
(NLP) submodular functions have been used for extractive text
summarization \cite{lin2012}.

To explain submodular functions, we introduce the following notation:
assume a finite set of data elements $V$, the {\em ground set}.  A valuation function $f:2^V
\rightarrow \mathbb{R}_+$ is then defined that returns a non-negative
real value for any subset $X \subseteq V$.  The function $f$ is called
\emph{submodular} if it satisfies the property of diminishing returns:
for all $X \subseteq Y$ and $v \notin Y$, the following is
true:\looseness-1
\begin{equation}
f(X \cup \{ v \}) - f(X) \geq f(Y \cup \{ v \}) - f(Y).
\end{equation}
This means that the incremental value (or gain) of element $v$
decreases when the context in which $v$ is considered grows from $X$
to $Y$. The ``gain'' is defined as $f(v | X) \triangleq f(X
\cup \{ v \}) - f(X)$. Thus, $f$ is submodular if $f(v|X) \geq f(v|Y)$.
Submodularity is a natural model for data selection in SMT and other
NLP tasks.  The ground set $V$ is the set of training data elements,
and elements are selected from this set according to a submodular
valuation function for any given subset of $V$.  The value of this
function diminishes for items that are (partially) redundant with
other items in the already-selected subset, which is precisely the
submodularity property. The specific function we utilize for the purpose
of MT data selection is as follows:
\begin{equation}
f(X) = \sum_{u \in U} w_u \phi_u (\sum_{x \in X}  m_u(x))
\label{eq:feature_based}  
\end{equation}
Here, $U$ is a set of features (such as words, n-grams, etc.), $X$ is a subset of $V$, $w$ is a non-negative weight, $\phi$ is a non-negative, non-decreasing concave function, and $m_u(x)$ is a score indicating how relevant $u$ is in sample $x$. Thanks to the concave function, the contribution of each feature $u$ in the context of an existing subset $X$ diminishes as $X$ grows. 

In our work the feature set $U$ consists of all n-grams up to a pre-specified length drawn from a representative in-domain data set. The feature relevance scores $m_u(x)$ are the tf-idf weighted counts of the the features (n-grams). The tf-idf (term frequency, inverse document frequency) values are computed by treating each sentence as a ``document''. That is, the weighting term is
\begin{equation}
 tf-idf(u) = c(u,x) * log \frac{|V|}{c(u,V)} 
\end{equation}
where $c(u,x)$ is the count of $u$ in $x$ (term frequency), and $c(u,V)$ is the number of sentences out of $V$ that $u$ occurs in. 

The above function can be optimized efficiently even for large data sets. Formally, we have the following optimization problem:
\begin{align}
X^* \in \argmax_{X \subseteq V, m(X) \leq b } f(X),
\label{eq:bm}
\end{align}
where $b$ is a known budget -- in the present context, the budget can be, e.g., the number of words or parallel sentences to select.  Solving this problem exactly is
NP-complete~\cite{feige1998threshold}, and
expressing it as an ILP procedure renders it impractical for large
data sizes.  When $f$ is submodular and the cost is just size
($m(X) = |X|$), then the simple greedy algorithm (detailed
in Algorithm 1) will have a {\bf worst-case guarantee} of $f(\tilde X^*) \geq
(1-1/e)f(X_{\text{opt}}) \approx 0.63 f(X_{\text{opt}})$ where
$X_{\text{opt}}$ is the optimal 
and $\tilde X^*$ is the greedy
solution \cite{nem78}.
\LinesNumbered
\begin{algorithm}[t]
{\bf Input:} Submodular function $f: 2^V \to \mathbb R_+$, cost vector $m$, budget $b$, finite set $V$. \\
{\bf Output:} $X_k$ where $k$ is the number of iterations. \\
Set $X_0 \gets \emptyset$ ;
$i \gets 0$ \;
\While{$m(X_i) < b$}{
  Choose $v_i$ as follows:
    $v_i \in \set{ \argmax_{v \in V \setminus X_i } \frac{f(\set{ v } | X_i)}{m(v)}  }$ \;
  $X_{i+1} \gets X_i \cup \set{ v_i }$ ;  $i \gets i + 1$ \;
}
\caption{The Greedy Algorithm}
\label{alg:greedy_for_polymatroid_max}
\end{algorithm}
This constant factor guarantee stays the same as $n$ grows; thus, it scales well to large data sets.
The application of this procedure to the selection of training data for large-scale SMT tasks was
described in \cite{kirch14}. Here, we apply it in the same way to the selection of out-of-domain data for a small-scale task.

\section{Data}
\label{sec:data}
The in-domain data available for the present study is the Transtac
corpus of Iraqi Arabic; the sizes of the training, tuning and
development test sets are shown in Table \ref{tab:transtac}.
\begin{table}
\begin{tabular}{|c|c|c|c|c|c|}\hline
 Train & Tune  & Dev & Test1 & Test2 & Test3 \\ \hline \hline
 7.6M & 64kk & 29k & 8k & 10k & 9k \\ \hline
\end{tabular}
\caption{Size of Iraqi Arabic Transtac corpus partitions (in words).}
\label{tab:transtac}
\end{table}
    
The out-of-domain data sources used for the selection experiments are listed in Table
\ref{tab:data}. We utilize 22 LDC corpora that include MSA and other
dialects of Arabic, notably Egyptian and Levantine. For example,
training corpora developed for the GALE, TIDES, and BOLT projects were
included, as were the Levantine Arabic Treebank, an Egyptian Arabic word
alignment corpus, and a corpus of dialectal Arabic web data (75\% Levantine, 25\% Egyptian)
that was translated through crowdsourcing (thus, translations are noisy).
Note that even though a corpus may be officially listed as MSA, it may contain segments of DA, especially when broadcast conversations (e.g., talkshows) are included.
\begin{table*}[htb]
\begin{tabular}{|l|l|l|c|r|} \hline
  LDC ID & Description & Genre & Dialect & Size \\ \hline \hline
  LDC2005E83 & GALE-Y1Q1 & BN, BC, WB & MSA & 170k \\ 
  LDC2006E34 & GALE-Y1Q2 & BC, WB & MSA & 126k  \\ 
  LDC2006E39 & Tides MT05 Eval & NW & MSA & 135k \\
  LDC2006E44 & Tides MT04 Eval & NW & MSA & 170k  \\
  LDC2006E85 & GALE-Y1Q3 & WB & MSA & 18k \\  
  LDC2006E92 & GALE-Y1Q4 & BN, WB & MSA & 293k \\
  LDC2012T06 & GALE-P2-BC & BC & MSA & 174k \\ 
  LDC2007E101 & GALE-P2R1 & BC, BN & MSA & 337k \\ 
  LDC2007E101 & GALE-P3R1 & BC, NW, BN & MSA & 530k \\ 
  LDC2007E46 & GALE-P2R2 & NW, WB & MSA & 87k \\ 
  LDC2007E87 & GALE-P2R3 & BC, BN, NW, WB & MSA & 188k \\ 
  LDC2008E40 & GALE-P3R2 & BC, BN & MSA & 268k  \\ 
  LDC2009E15 & GALE-P4R1v2 & WB, BC, BN, NW & MSA & 305k  \\ 
  LDC2009E16 & GALE-P4R2 & BC, BN, NW, WB & MSA & 273k \\ 
  LDC2009E95 & GALE-P4R3v1.2 & BC, BN, NW, WB & MSA & 147k \\ 
  LDC2010E38 & GALE-P3 Treebank & BC, NW  & MSA & 349k\\ 
  LDC2010E79 & GALE-Levantine & BC & Levantine & 34k \\ 
  LDC2010T17 & NIST-OpenMT-2006 &  NW, BC, BN, WB & MSA & 141k \\ 
  LDC2010T23 & NIST-OpenMT-2009 &  NW, WB & MSA & 129k \\ 
  LDC2012E19 & BOLT-P1-R2 MT Training Data & DF & Egyptian & 126k\\ 
  LDC2012E51 & BOLT-P1 ARZ word alignments & DF & Egyptian & 55k \\
  LDC2012T09 & Web translations & Various & Egyptian, Levantine & 1.613M \\ \hline
  \multicolumn{4}{|c|}{{\bf Total}} & 5.690M \\ \hline
\end{tabular}
\caption{List of out-of-domain corpora used for data selection. BN = broadcast news, BC = broadcast conversations, WB = web blogs, NW = newswire data, DF = discussion forums. Sizes are given in number of source-language words after tokenization.}
\label{tab:data}
\end{table*}

\section{Experiments and Results}
\label{sec:exp}

We use two different MT systems for translation from IA to English, an
in-house system based on Moses and the SRI MT system developed for the
DARPA BOLT (Broad Operational Language Translation) spoken dialog
translation project (see \cite{ayan13,kirch15} for more details).  The
former is a flat phrase-based statistical MT system with a
hierarchical lexicalized reordering model and a 6-gram language model
trained on the target side of the Transtac training data. For
preprocessing we use a statistical morphological segmenter developed
in the BOLT project. The second system is similar in nature but has a
hierarchical phrase-based translation model and utilizes sparse
features (see \cite{zhao14} for more information).

\subsection{Initial evaluation of selection techniques}

In an initial set of experiments we attempted to gauge the
performance of the cross-entropy vs.~the submodular selection technique
by subselecting the Transtac training data. We chose 10-40\% of the Transtac training
set; the feature set $U$ was the set of all n-grams up to length 7 of the tune and dev sets.
We investigated both translation directions, IA $\rightarrow$ English and English $\rightarrow$ IA.
Table \ref{tab:select} shows the BLEU scores.
\begin{table}
\begin{tabular}{|l||c|c|c|c|} \hline
 & \multicolumn{2}{c}{IA $\rightarrow$ EN} & \multicolumn{2}{|c|}{EN $\rightarrow$ IA} \\ \hline
Size & Xent & SM & Xent & SM \\ \hline \hline
 10\% & 30.2 & {\bf 32.3} & 16.1 & {\bf 17.9}\\ \hline
 20\% & 31.5 & {\bf 32.5} & 17.0 & {\bf 17.7}\\ \hline
 30\% & 31.8 & {\bf 32.5} & 17.4 & {\bf 17.5}\\ \hline
 40\% & 31.2 & {\bf 32.6} & 17.3 & {\bf 17.5}\\ \hline
 100\% & \multicolumn{2}{c}{32.5} & \multicolumn{2}{|c|}{16.2} \\ \hline
\end{tabular}
\caption{BLEU scores on dev set for training data subselection using cross-entropy (Xent) vs.~the submodular (SM) method. IA = Iraqi Arabic, EN = English.}
\label{tab:select}
\end{table}

Compared to using 100\% of the training data, the same or even better
performance can be obtained by using a subset of the data when the
submodular subselection technique is used, even at small percentages
of the training data.  The cross-entropy method falls short of this
performance, presumably due to the failure of this method to control
for redundancy in the selected set.

\subsection{Selection of out-of-domain training data}

In order to integrate additional out-of-domain training data, we set a
budget constraint of 100k words on the source side. The LDC corpora
were preprocessed in the same manner as the Transtac data, i.e.~, they
were preprocessed and morphologically segmented. The greedy algorithm
was used in combination with Equation \ref{eq:feature_based} to select
parallel sentences from the corpora listed in Table \ref{tab:data}
such that the resulting corpus contains at most 100k words on the
source side. The selected data was then added to both the MT and LM
training data. Table \ref{tab:ood1} shows the BLEU scores and
position-independent word error rate (PER) for the in-house MT system that
was used for development purposes (note that baseline results are
different from those in Table \ref{tab:select} because the baseline MT system
changed in between experiments and was trained on different data set definitions and tokenization schemes).
We again compared the cross-entropy 
against the submodular selection method. Improvements in the system
are small; however, the submodular technique again shows slightly
better results.

\begin{table}
\begin{tabular}{|l|c|c|} \hline
          & \multicolumn{2}{|c|}{IA-EN}\\  \hline
         & BLEU (\%) & PER (\%) \\ \hline
         Baseline & 33.5 & 40.9 \\ \hline
         Xent & 33.6 & 40.9\\ \hline  \hline
         Submod  & {\bf 33.7} & {\bf 40.7}\\ \hline 
          & \multicolumn{2}{|c|}{EN-IA}\\  \hline
          & BLEU (\%) & PER (\%) \\ \hline
          Baseline & 17.0 & 57.1 \\ \hline
          Xent & 17.1 & 57.1 \\ \hline \hline
          Submod & {\bf 17.2} & {\bf 56.8} \\ \hline
        \end{tabular}
\caption{BLEU and PER on dev set for system with additional out-of-domain data, in-house system.}
\label{tab:ood1}
\end{table}

We subsequently used the selected data with the submodular method in
the second MT system, viz.~the evaluation system developed for a
bilingual dialog system, and tested the system on  additional
in-domain data sets.  BLEU scores (shown in Table \ref{tab:eval}) show
slight improvements of up to 0.5 absolute. Note that the selected data set
was very small, containing only 7k sentences. Larger sets (up to 20k) were tried but
were not found to be useful. 
  
\begin{table}        
\begin{tabular}{|l||c||c|c|c|} \hline
System  & dev & test1 & test2 & test3 \\ \hline
base & 17.5 & 35.2 & 32.2 & 33.1 \\ \hline
+ 7k data & 17.5 & {\bf 35.5} & {\bf 32.7} & {\bf 33.5} \\ \hline
\end{tabular}
\caption{BLEU scores of EN-IA system, obtained with an additional 7k sentences of submodular-selected data, evaluation system.}
\label{tab:eval}
\end{table}

We analyzed the selected data as to its origin and found that the top
three data sources were broadcast conversations from various GALE
corpora (47\%), the dialectal web corpus (35.7\%), and the BOLT MT
training data (9.9\%). 

\subsection{Using translated speech data}

In addition to the various parallel text corpora listed in Table \ref{tab:data} we also had
access to an Iraqi Arabic Conversational Telephone Speech (CTS) corpus (LDC2006T16). This corpus includes
with speech transcriptions but no translations. Although the data matches the dialect of interest
is is not necessarily matched in topic or style. To obtain parallel data we translated 
the transcriptions of this corpus with our baseline IA $\rightarrow$ EN translation system.
Those segments that were translated contiguously (i.e., without intervening out-of-vocabulary
words) were extracted and added to the data from the corpora in Table \ref{tab:data}.
Data selection was then re-run. We found that in this experiment 80\% of the selected
data came from the CTS corpus; however, the translation performance did not improve
(see Table \ref{tab:cts}). The likely reason is that translations were too noisy to be
used as parallel data and introduced more confusability and irrelevant variation rather than
contributing useful translations. The use of automatically translated speech data might be improved
by selecting only the most confident translations according to the translation model scores.

\begin{table}
\begin{tabular}{|l|c|c|} \hline
          & \multicolumn{2}{|c|}{IA-EN}\\  \hline
         & BLEU (\%) & PER (\%) \\ \hline
         Baseline & 33.5 & 40.9 \\ \hline
         Submod  & {\bf 33.7} & {\bf 40.7}\\ \hline 
          + CTS & 33.8 & 41.0 \\ \hline \hline
          & \multicolumn{2}{|c|}{EN-IA}\\  \hline
          & BLEU (\%) & PER (\%) \\ \hline
          Baseline & 17.0 & 57.1 \\ \hline
          Submod & {\bf 17.2} & {\bf 56.8} \\ \hline
          + CTS & 17.0 & 57.5 \\ \hline
        \end{tabular}
\caption{BLEU and PER on dev set, system with additional out-of-domain data, including CTS, in-house system.}
\label{tab:cts}
\end{table}

\section{Conclusion}
\label{sec:conc}

We have described data selection procedures for identifying Iraqi
Arabic data resources in unrelated dialectal and/or MSA corpora. We
have demonstrated that judiciously selected data can improve MT performance
even when the overall amount is very small. Furthermore, we have
compared two different data selection techniques, the widely-used
cross-entropy selection method, and a more recently developed method
that relies on submodular function optimization. The latter performed
slightly better than the former. Finally, we have conducted initial
experiments on utilizing automatically translated conversational
speech as additional training data. Whereas the data was strongly
matched to the in-domain data on the source side, the translations
were too noisy to yield any further improvement in machine translation
performance.\\
\noindent
\begin{small}
{\bf Acknowledgments}\\
This study was funded by the Defense Advanced Research Projects Agency (DARPA)
under contract HR0011-12-C-0016 - subcontract 19-000234 and by the Intelligence Advanced
Research Projects Activity (IARPA) under agreement number FA8650-12-2-7263.
The U.S. Government is authorized to reproduce and
distribute reprints for Governmental purposes notwithstanding any
copyright notation thereon. The views and conclusions contained herein
are those of the authors and should not be interpreted as necessarily
representing the official policies or endorsements, either expressed
or implied, of Intelligence Advanced Research Projects Activity
(IARPA) or the U.S.˜Government.
\end{small}
\bibliographystyle{eamt12}
\bibliography{bolt-arXiv} 
\end{document}